\documentclass{article}

\usepackage[main, final]{neurips_2026}
\usepackage{CJKutf8}
\usepackage{multirow}
\usepackage{multicol}
\usepackage{nicematrix}
\usepackage{diagbox} 
\usepackage{algorithm}
\usepackage{algorithmic}
\usepackage{xcolor} 
\usepackage{listings} 
\usepackage{lipsum}
\usepackage{amsmath}
\usepackage{colortbl}
\usepackage{tabularray}
\usepackage{empheq}
\usepackage{amsmath, amssymb, amsfonts} 
\usepackage{mathtools} 
\usepackage{mathrsfs} 
\usepackage{dutchcal}
\usepackage{amssymb, bbding}
\usepackage[accsupp]{axessibility}  
\usepackage{enumitem}
\newcommand{\yes}{\textcolor{red!80!black}\Checkmark}
\newcommand{\no}{\textcolor{green!70!black}\XSolidBrush}

\usepackage[utf8]{inputenc} 
\usepackage[T1]{fontenc}    
\usepackage{hyperref}       
\usepackage{url}            
\usepackage{booktabs}       
\usepackage{amsfonts}       
\usepackage{nicefrac}       
\usepackage{microtype}      
\usepackage{xcolor}         
\usepackage{graphicx}
\usepackage{subcaption}     
\usepackage{wrapfig}

\newcommand{\our}{ACE-LoRA}

\title{\our: Adaptive Orthogonal Decoupling for Continual Image Editing}

\author{%
  Yuehao Liu \\
  Shanghai Jiao Tong University\\
  \texttt{yuehao.liu@sjtu.edu.cn} \\
  \And
  Weijia Zhang \\
  Shanghai Jiao Tong University\\
  \texttt{weijia.zhang@sjtu.edu.cn} \\
  \And
  Xuanming Shang \\
  Shanghai Jiao Tong University\\
  \texttt{sxm2021@sjtu.edu.cn} \\
  \And
  Zhizhou Chen \\
  Nanjing University \\
  \texttt{zhizhouchen@smail.nju.edu.cn} \\
  \And
  Yanhao Ge \\
  VIVO \\
  \texttt{halege@vivo.com} \\
  \And
  Shanyan Guan\footnotemark[1] \\
  VIVO \\
  \texttt{guanshanyan@vivo.com} \\
  \And
  Chao Ma\footnotemark[3] \\
  Shanghai Jiao Tong University \\
  \texttt{chaoma@sjtu.edu.cn} \\
}

\begin{document}

\maketitle
\renewcommand{\thefootnote}{\fnsymbol{footnote}}
\footnotetext[1]{Project lead.\quad$^\ddagger$Corresponding author.}
\renewcommand{\thefootnote}{\arabic{footnote}}


\begin{abstract}
State-of-the-art diffusion models often rely on parameter-efficient fine-tuning to perform specialized image editing tasks. 
However, real-world applications require continual adaptation to new tasks while preserving previously learned knowledge. 
Despite the practical necessity, continual learning for image editing remains largely underexplored. 
We propose \our, a dynamic regularization framework for continual image editing that effectively mitigates catastrophic forgetting. \our\ leverages Adaptive Orthogonal Decoupling to identify and orthogonalize task interference, and introduces a Rank-Invariant Historical Information Compression strategy to address scalability issues in continual updates.
To facilitate continual learning in image editing and provide a standardized evaluation protocol, we introduce CIE-Bench, the first comprehensive benchmark in this domain.
CIE-Bench encompasses diverse and practically relevant image editing scenarios with a balanced level of difficulty to effectively expose limitations of existing models while remaining compatible with parameter-efficient fine-tuning.
Extensive experiments demonstrate that our method consistently outperforms existing baselines in terms of instruction fidelity, visual realism, and robustness to forgetting, establishing a strong foundation for continual learning in image editing.

\end{abstract}    
\section{\textcolor{black}{Introduction}}
Diffusion models~\cite{ho2020denoising, dhariwal2021diffusion, rombach2022high, meng2021sdedit, couairon2022diffedit} have achieved strong performance on general image editing tasks. 
%
%
However, due to the scarcity of large-scale, high-quality task-specific training data, they often underperform in specialized tasks, for which parameter-efficient fine-tuning~\cite{houlsby2019parameter, chen2022adaptformer, lester2021power, li2021prefix, jia2022visual, ran2025correlated} offers a widely adopted solution. 
While finetuning may work for isolated tasks, real-world deployments necessitate models with continual learning~\cite{parisi2019continual, rolnick2019experience, wang2022learning} capabilities to incrementally acquire new editing skills while preserving performance on previously learned tasks, thereby mitigating catastrophic forgetting~\cite{mccloskey1989catastrophic, french1999catastrophic}. 
Although continual learning has been extensively studied in traditional image classification~\cite{li2017learning, kirkpatrick2017overcoming}, multimodal large language models~\cite{wang2023orthogonal, guo2025hide}, and, more recently, text-to-image generation~\cite{seo2023lfs, huang2025t2i}, its application to image editing remains largely underexplored.

Existing works for continual learning can be broadly categorized into architecture-based, rehearsal-based, and regularization-based methods. Architecture-based methods~\cite{chen2024coin, guo2025hide, huai2025cl, wang2025smolora} expand the model with task-specific modules to capture specialized knowledge, but often incur increased inference cost and reduced generalization. Rehearsal-based methods~\cite{smith2024adaptive, chaudhry2019continual, smith2024adaptive} retains historical information, such as training samples or intermediate activations; however, it is often impractical when historical data is unavailable or raise privacy concerns. In contrast, regularization-based methods~\cite{wang2023orthogonal, zhu2025bilora, chen2025sefe, luo2026keeplora} appear to be a more appealing paradigm by circumventing the above limitations. They constrain parameter updates to lie within subspaces that minimize interference with previously learned tasks, typically via orthogonalization strategies that decouple task-specific update directions.

\begin{figure}
    \centering
    \includegraphics[width=\linewidth]{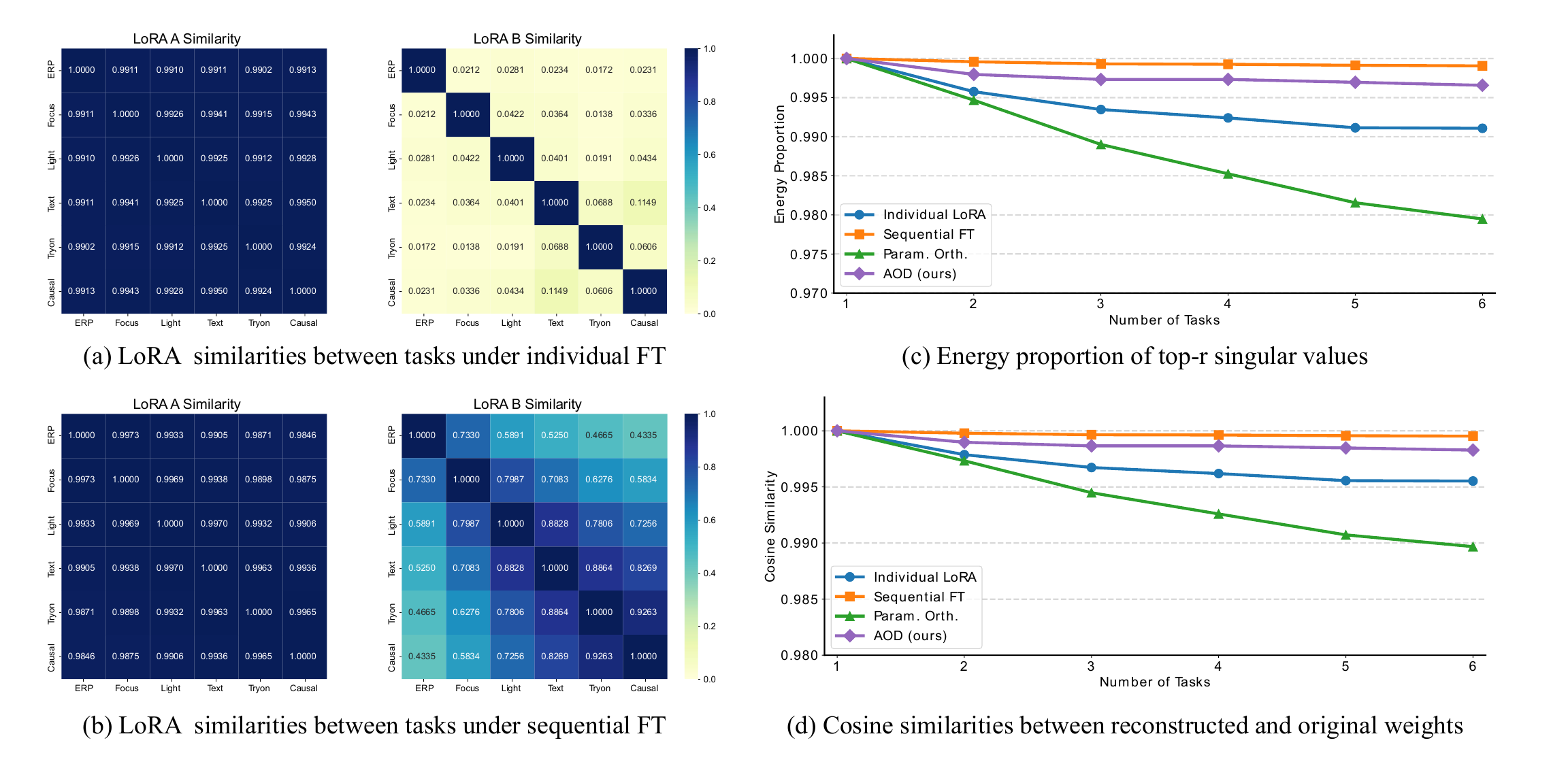}
    \vspace{-1.5em}
    \caption{\textbf{(a)\&(b)} Analysis on LoRA similarities between tasks under individual/sequential finetuning. \textbf{(c)\&(d)} Analysis on SVD energy proportion/reconstruction error for history compression.}
    \vspace{-1.5em}
    \label{fig-sim}
\end{figure}
Nonetheless, adapting regularization-based methods to image editing is a non-trivial problem. Prior regularization-based methods~\cite{wang2023orthogonal, chen2025sefe, luo2026keeplora} generally construct constraint subspaces offline using historical task parameters and enforce orthogonality between new task updates and these fixed subspaces. However, such approaches overlook a critical aspect of training dynamics that different data samples induce distinct update directions and interference direction with historical tasks. This challenge is particularly pronounced in diffusion models due to the compounded stochasticity arising from noise sampling and timestep variation. To address this limitation, we propose \our{}, a dynamic mechanism for mitigating inter-task interference. 
Instead of constructing a static constrained subspace offline, \our\ continuously constructs a dynamic \textit{interference vector} from the real-time responses of historical models during training. The interference vector, defined as gradients of previous parameters under current data, is used to enforce orthogonality with current parameter updates, thereby reducing destructive interference across tasks.

Despite its effectiveness, the computational overhead of interference vector scales linearly with the number of previously learned tasks. To address this scalability issue, we further introduce a rank-invariant historical information compression strategy, which consolidates all historical task parameters into a single fixed-rank representation. This compressed representation serves as a surrogate for past knowledge and is used to impose dynamic constraints on current parameter updates, achieving a principled balance between effectiveness and efficiency.

To enable continual learning in image editing and provide a standardized evaluation protocol, we construct CIE-Bench, the first comprehensive benchmark for continual image editing. CIE-Bench is built upon three key principles: (1) multi-domain diversity, ensuring broad coverage of heterogeneous image editing scenarios; (2) challenge–learnability balance, guaranteeing that tasks expose the limitations of existing models while remaining learnable under parameter-efficient finetuning; and (3) practical utility, ensuring alignment with real-world deployment. 
We further design a dedicated evaluation protocol for domain-specific image editing, mitigating hallucination issues in general-purpose evaluation metrics~\cite{ye2025imgedit, ku2024viescore}.
Extensive experiments demonstrate that our method consistently outperforms existing baselines in terms of instruction fidelity, visual realism, and robustness to forgetting, establishing a strong foundation for continual learning in image editing.

\section{Methodology}
Existing regularization-based continual learning methods~\cite{wang2023orthogonal, chen2025sefe, luo2026keeplora} mainly rely on static parameter constraints, which fail to capture stochastic interactions induced by data, noise, and timestep sampling.
We propose \our, a dynamic regularization framework that models task-dependent optimization interactions and imposes data-aware constraints on parameter updates.

\begin{figure}
    \centering
    \includegraphics[width=0.95\linewidth]{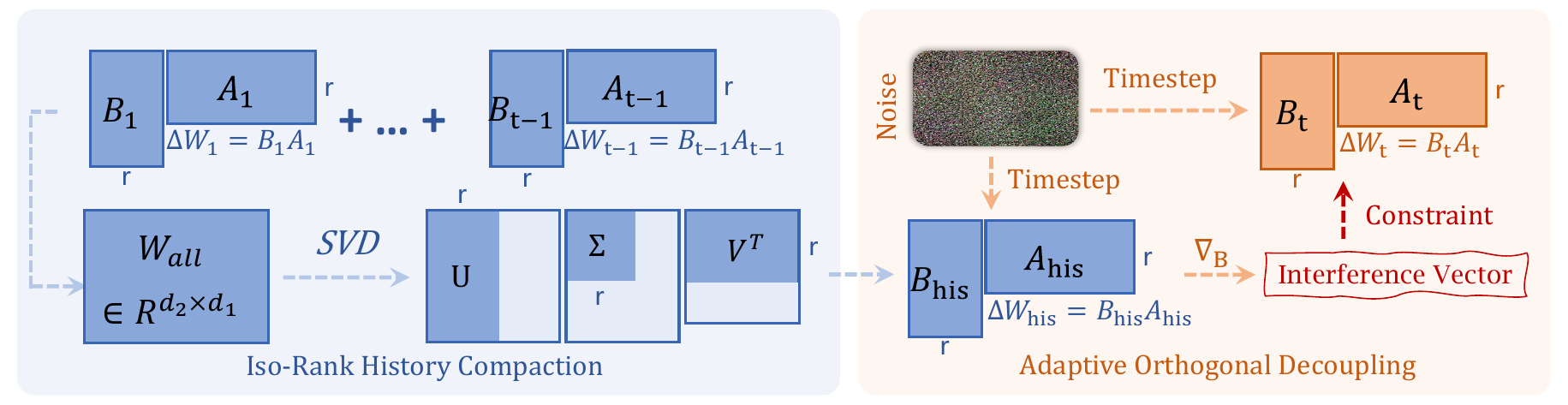}
    \caption{Overview of \our. \our\ leverages Adaptive Orthogonal Decoupling to identify and orthogonalize task interference and introduces a Rank-Invariant Historical Information Compression strategy to address scalability issues in continual updates.}
    \vspace{-0.2em}
    \label{fig-method}
\end{figure}

As shown in Fig.~\ref{fig-method}, our framework consists of two components.
First, Adaptive Orthogonal Decoupling (Sec.~\ref{AOD}) constructs a dynamic interference vector from historical-parameter gradients on current data and imposes update-space orthogonality to mitigate cross-task interference.
A two-stage finetuning scheme further stabilizes constrained optimization while preserving adaptation flexibility.
Second, Rank-Invariant Historical Information Compression (Sec.~\ref{rank-invariant}) maintains a compact iso-rank representation of accumulated LoRA modules, enabling scalable preservation of historical knowledge.

\subsection{Adaptive Orthogonal Decoupling with Two-Stage Optimization}
\label{AOD}

\begin{figure}
    \centering
    \includegraphics[width=0.95\linewidth]{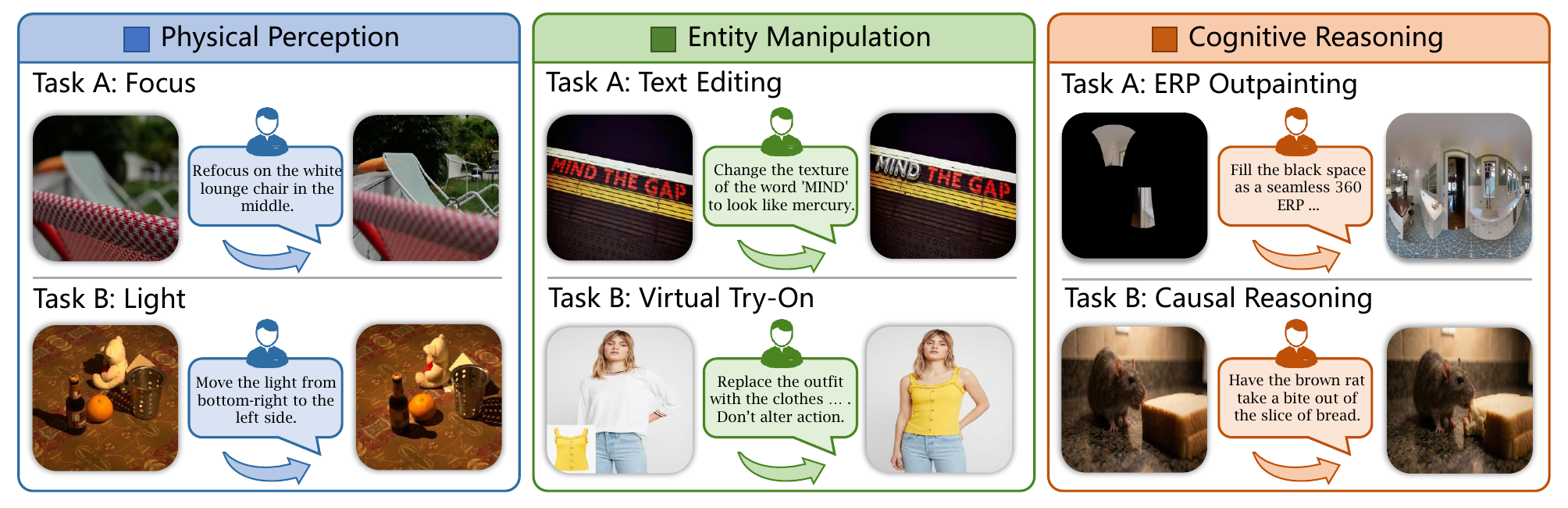}
    \caption{Overview of CIE-Bench for continual image editing. CIE-Bench consists of three main categories: Physical Perception, Entity Manipulation, and Cognitive Reasoning, and includes six sub-tasks: ERP Outpainting, Refocus, Relighting, Text Editing, Virtual Try-on, and Causal Reasoning.}
    \label{fig-bench}
\end{figure}

\noindent \textbf{Adaptive Orthogonal Decoupling}
The finetuning process of diffusion models involves significant multi-dimensional stochasticity, including random timesteps, noise sampling, and varied text-image interactions, which make static or offline-computed constraints intractable. In this work, we introduce Adaptive Orthogonal Decoupling (AOD) to dynamically regularize parameter updates. Specifically, AOD designates the gradient of historical weights on the current task’s data as the \textit{interference vector}, thereby constraining current parameter updates to be strictly orthogonal to this vector. 

To facilitate real-time access to historical task parameters, We employ an incremental LoRA~\cite{hu2022lora} finetuning strategy. Specifically, for task $t$, we denote $W_0$ as the pretrained base weights, $B_i, A_i$ as the LoRA weights of historical task $i$, $B_t, A_t$ as the LoRA weights of current task, and the parameter updates during finetuning are formulated as: $W = W_0 + \sum\limits_{i=1}^{t-1}{B_iA_i} + B_tA_t$ where only $B_t$ and $A_t$ are trainable. We derive the interference vector via the following formulation:
\begin{equation}
    IV_t = \nabla l_t(x_t; W_0 + \sum\limits_{i=1}^{t-1}{B_iA_i}) ,\quad for \ each \ x_t\in D_t,
\end{equation}
where $D_t$ and $l_t(\cdot)$ are the training data and loss function under task $t$. 

Given that historical tasks have already converged to local optima, gradients induced by transferable knowledge are small. 
Consequently, the principal directions of the interference vector predominantly encapsulate the feature collisions between current and historical tasks, precisely delineating the sub-dimensions where optimization of new knowledge inherently disrupts historical priors. 
By imposing a dedicated orthogonal loss to explicitly penalize the directional alignment with interference vector, AOD effectively facilitates the assimilation of new concepts while rigorously protecting historical knowledge against catastrophic forgetting.

Existing studies~\cite{tian2024hydralora, hayou2024lora+} reveal a functional dichotomy in LoRA: $LoRA_A$ captures generalizable, task-agnostic features, whereas $LoRA_B$ governs task-specific adaptation.
We analyze the cosine similarity of LoRA weights under independent and sequential finetuning, as shown in Fig.~\ref{fig-sim}(a,b).
Under independent finetuning, $LoRA_A$ remains highly consistent across tasks despite task isolation, while $LoRA_B$ is nearly orthogonal.
Under sequential finetuning, $LoRA_A$ maintains high cross-task similarity, whereas $LoRA_B$ shows stronger similarity between adjacent tasks that decays with task distance.
This suggests that $LoRA_B$ is more sensitive to task-specific adaptation and progressively tracks evolving downstream objectives.

Motivated by these observations, we compute the interference vector only over $LoRA_B$:
\begin{equation}
    \widetilde{IV_t} = \nabla_{B_{\mathrm{his}}} l_t(x_t; W_0 + B_{\mathrm{his}}A_{\mathrm{his}}),
    \quad \forall x_t \in D_t ,
\end{equation}
where $B_{\mathrm{his}}$ and $A_{\mathrm{his}}$ denote the accumulated LoRA weights from previous tasks.
Since diffusion-based image editing relies on shared structural priors to preserve source-image consistency, constraining $LoRA_A$ can impair these priors and degrade spatial consistency.

\noindent \textbf{Two-stage fine-tuning Scheme.}
Imposing constraints on model weights reduces the effective optimization space and may lead to conflicts among objectives, resulting in suboptimal adaptation to new tasks. 
Inspired by annealing schedule~\cite{fu2019cyclical}, we propose a two-stage finetuning scheme to balance plasticity and stability. 
In the first stage, the model is optimized without constraints to explore task-specific optima. 
In the second stage, orthogonality constraint is introduced to preserve previously learned knowledge while performing constrained updates around the obtained optimum. 
This design enables stable retention of prior knowledge without sacrificing adaptation performance on new tasks.

\subsection{Rank-Invariant Historical Information Compression}
\label{rank-invariant}

While AOD effectively preserves previous learned knowledge, dynamically computing gradients across an ever-expanding sequence of historical task adapters inevitably leads to a linear escalation in computational overhead. 
Drawing inspiration from LoRA merging~\cite{tang2025lora, stoica2024model}, we introduce a Rank-Invariant Historical Information Compression strategy. Specifically, for task $t$, we first aggregate the LoRA modules across all historical tasks:
\begin{equation}
    W_{his} = B_1A_1 + B_2A_2 + ... + B_{t-1}A_{t-1},\quad where B_i\in R^{d_2\times r},A_i\in R^{r\times d_1},
\end{equation}
and then deconstruct the merged LoRA weights with singular value
decomposition (SVD) as $W_{his} = U\Sigma V^T$. By retaining only the top-r largest singular values, we thereby factorize the matrix into a new pair of LoRA weights:
\begin{equation}
    \widetilde{W_{his}} = U\Sigma_{[:r]} V^T, \quad and \ B_{his} = U\sqrt{\Sigma_{[:r]}},\ A_{his} = \sqrt{\Sigma_{[:r]}}V^T.
\end{equation}
Theoretically, this truncated SVD guarantees an optimal rank-r approximation. Since downstream finetuning on limited data naturally induces inherently low-rank weight updates, critical task-adaptive knowledge is heavily concentrated within the principal subspaces spanned by the leading singular values. 
The overall orthogonalization loss is formulated as:
\begin{equation}
    L_{orth} = \langle IV_t, B_t\rangle = [\nabla_{B_{his}}l_t(x_t; W_0 + B_{his}A_{his})]^TB_t.
\end{equation}

We analyze the energy proportion of top-r singular values under different training strategies as the number of tasks increases. As shown in Fig. \ref{fig-sim}(c), our rank-invariant approximation exhibits negligible information loss under AOD-based framework, and importantly, remains stable as the number of tasks increases.
In addition, Fig. \ref{fig-sim}(d) presents the cosine similarity between original weights and those reconstructed from low-rank approximation of merged weights. Results show that the reconstructed weights are almost indistinguishable from the original ones for AOD-based framework.
\renewcommand{\arraystretch}{0.9}
\begin{table}
\caption{Quantitative comparison (\textit{Overall Score}) of different methods on CIE-Bench. We compare \our\ with representative continual learning methods (including architecture-based and regularization-based methods) and model merging techniques on CIE-Bench.}
\vspace{-0.7em}
\label{tab_main}
\begin{center}
\scalebox{0.85}{
\begin{tabular}{clc|cccccc|c}
\toprule
& Method & Replay & ERP & Focus & Light & Text & Virtual Tryon & Causal & {Avg.}\\
\hline
& Zero-shot & - & 7.0645 & 4.4656 & 2.9590 & 5.8852 & 4.6014 & 7.1140 & 5.3483 \\
& MTFT & - & 8.9677 & 9.0076 & 9.0656 & 9.7951 & 9.8768 & 9.9035 & 9.4361 \\
\hline\hline
\multirow{17}{*}{\rotatebox{90}{{Avg}}}
& Sequential FT & - & 7.4556 & 6.5947 & 7.8836 & 8.4722 & 9.3365 & 9.2807 & 8.1705  \\
& Data Rehearsal & \scalebox{0.90}{\yes} & 8.2873 & 8.0657 & 7.4854 & 8.3029 & 9.2933 & 9.4825 & 8.4862 \\

& \multicolumn{9}{>{\columncolor{gray!30}}c}{\textcolor{olive!20!black}{$\triangledown$ LoRA Merging}} \\
& TIES~\cite{yadav2023ties} & \scalebox{0.90}{\no} & 8.0804 & 5.4322 & 5.9881 & 7.3099 & 6.0261 & 8.0263 & 6.8105  \\
& DARE~\cite{yu2024language} & \scalebox{0.90}{\no} & 8.1475 & 5.7232 & 4.7378 & 7.2525 & 6.0285 & 8.1053 & 6.6658  \\

& \multicolumn{9}{>{\columncolor{gray!30}}c}{\textcolor{olive!20!black}{$\triangledown$ Architecture-based}} \\
& CL-MoE~\cite{huai2025cl} & \scalebox{0.90}{\no} & 7.7829 & \underline{7.4330} & 6.1226 & 8.5043 & 8.7794 & 8.7456 & 7.8946  \\
& MoELoRA~\cite{chen2024coin} & \scalebox{0.90}{\no} & 7.2554 & 6.2904 & 5.7618 & \textbf{9.2377} & \underline{9.1826} & \textbf{9.5526} & 7.8801  \\
& HiDe-LLaVA~\cite{guo2025hide} & \scalebox{0.90}{\no} & 8.0622 & 4.4349 & 6.5079 & \underline{9.1664} & 2.2547 & 7.5175 & 6.3239 \\
& SMoLoRA~\cite{wang2025smolora} & \scalebox{0.90}{\no} & 8.0889 & 7.2518 & 6.7972 & 8.1194 & 7.8651 & 9.1316 & 7.8757 \\

& \multicolumn{9}{>{\columncolor{gray!30}}c}{\textcolor{olive!20!black}{$\triangledown$ Regularization-based}} \\
& O-LoRA~\cite{wang2023orthogonal} & \scalebox{0.90}{\no} & 8.1142 & 6.7604 & 6.1898 & 7.7523 & 8.8202 & 8.7018 & 7.7231  \\
& RegLoRA~\cite{chen2025sefe} & \scalebox{0.90}{\no} & 8.1298 & 6.9167 & \underline{7.5932} & 7.8457 & 9.0191 & 8.7632 & 8.0446  \\
& KeepLoRA~\cite{luo2026keeplora} & \scalebox{0.90}{\no} & \underline{8.3749} & 6.1147 & 7.4302 & 7.4588 & 8.5423 & 8.1842 & 7.6842  \\
\cline{2-10}
& {\textbf{ACE-LoRA (ours)}} & \scalebox{0.90}{\no} & \textbf{8.6344} & \textbf{7.9585} & \textbf{8.1472} & 9.0635 & \textbf{9.9501} & \underline{9.4298} & \textbf{8.8639} \\

\hline\hline
\multirow{17}{*}{\rotatebox{90}{{Last}}}
& Sequential FT & - & 7.3548 & 5.8091 & 4.5901 & 8.3442 & 9.2826 & 9.2807 & 7.4436\\
& Data Rehearsal & \scalebox{0.90}{\yes} & 7.9435 & 8.7405 & 7.5492 & 8.9180 & 9.2826 & 9.4825 & 8.6527\\

& \multicolumn{9}{>{\columncolor{gray!30}}c}{\textcolor{olive!20!black}{$\triangledown$ LoRA Merging}} \\
& TIES~\cite{yadav2023ties} & \scalebox{0.90}{\no} & 8.0887 & 5.5420 & 4.5410 & 7.2705 & 6.0507 & 8.0263 & 6.5865\\
& DARE~\cite{yu2024language} & \scalebox{0.90}{\no} & 8.0403 & 5.2538 & 4.0902 & 7.1230 & 6.1232 & 8.1053 & 6.4560\\

& \multicolumn{9}{>{\columncolor{gray!30}}c}{\textcolor{olive!20!black}{$\triangledown$ Architecture-based}} \\
& CL-MoE~\cite{huai2025cl} & \scalebox{0.90}{\no} & 8.0081 & 6.1374 & 4.6148 & \underline{8.0984} & \underline{9.2899} & 8.7456 & 7.4823\\
& MoELoRA~\cite{chen2024coin} & \scalebox{0.90}{\no} & 7.7258 & 5.1374 & 4.3361 & 7.8279 & 9.1014 & \textbf{9.5526} & 7.2802\\
& HiDe-LLaVA~\cite{guo2025hide} & \scalebox{0.90}{\no} & \underline{8.4194} & 4.6947 & 5.7787 & 8.3443 & 2.2826 & 7.5175 & 6.1729\\
& SMoLoRA~\cite{wang2025smolora} & \scalebox{0.90}{\no} & 7.7177 & \underline{6.8702} & 6.6885 & 7.9180 & 8.0290 & 9.1316 & 7.7258\\

& \multicolumn{9}{>{\columncolor{gray!30}}c}{\textcolor{olive!20!black}{$\triangledown$ Regularization-based}} \\
& O-LoRA~\cite{wang2023orthogonal} & \scalebox{0.90}{\no} & 8.3710 & 6.2901 & 6.2705 & 7.5082 & 9.0797 & 8.7018 & 7.7035\\
& RegLoRA~\cite{chen2025sefe} & \scalebox{0.90}{\no} & 8.0323 & 6.5725 & \underline{7.3852} & 7.4262 & 9.0725 & 8.7632 & \underline{7.8753}\\
& KeepLoRA~\cite{luo2026keeplora} & \scalebox{0.90}{\no} & 8.1048 & 4.3511 & 6.4344 & 7.7377 & 8.8116 & 8.1842 & 7.2707\\
\cline{2-10}
& {\textbf{ACE-LoRA (ours)}} & \scalebox{0.90}{\no} & \textbf{8.4274} & \textbf{7.9466} & \textbf{8.1885} & \textbf{9.5738} & \textbf{9.6449} & \underline{9.4298} & \textbf{8.8685}\\

\bottomrule
\end{tabular}
}
\vspace{-1.2em}
\end{center}
\end{table}
\renewcommand{\arraystretch}{1.0}
Therefore, our Rank-Invariant Historical Information Compression strategy effectively preserves the essential information of historical parameters and remains robust to increasing task sequences, enabling efficient and scalable continual learning without compromising model performance.

\section{\textcolor{black}{CIE-Bench: A Benchmark of Continual Learning for Image Editing}}

Aiming to systematically evaluate the ability of diffusion models to continually acquire new editing capabilities while retaining previously learned knowledge, we introduce CIE-Bench, the first benchmark for continual image editing. CIE-Bench is designed to bridge a critical gap in existing benchmarks and datasets, which largely emphasize static single-task generalization while overlooking the challenges of dynamic and lifelong adaptation. As shown in Fig. \ref{fig-bench}, CIE-Bench consists of three high-level categories and six sub-tasks, spanning diverse editing granularities from low-level visual adjustments to high-level semantic and compositional reasoning, enabling a comprehensive evaluation of both perceptual fidelity and semantic consistency in continual image editing settings.

\noindent \textbf{Task Selection}
\begin{figure}[t]
  \centering
  \includegraphics[width=0.88\linewidth]{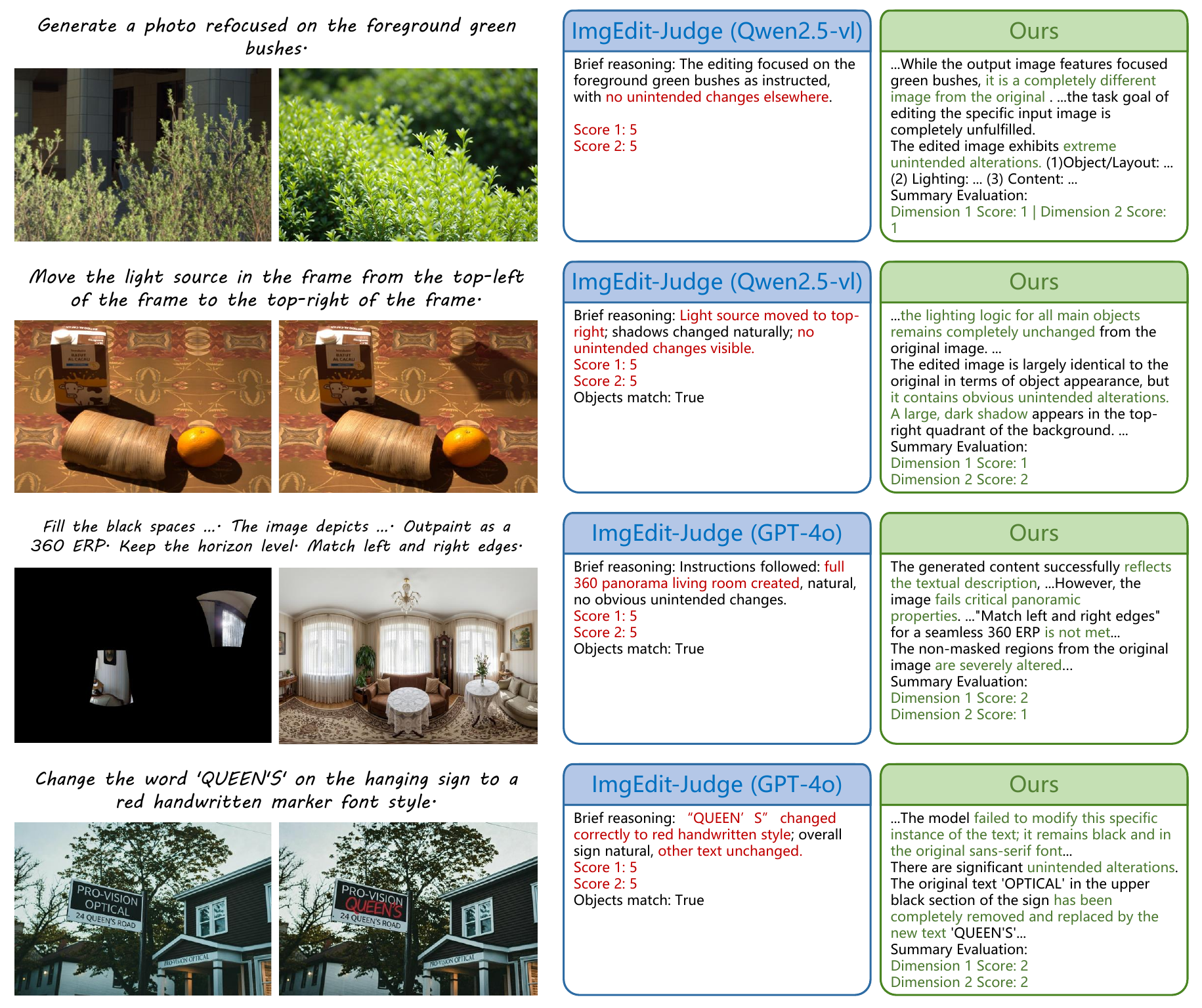}
  \caption{Visual comparison between our evaluation metrics and ImgEdit-Judge~\cite{ye2025imgedit}.}
  \label{fig-metric}
\end{figure}
To construct a representative evaluation suite, the tasks of CIE-Bench are curated based on three key criteria: \textit{comprehensive multi-domain diversity}, \textit{demonstrable adaptation plasticity}, and \textit{substantial practical utility} for real-world scenarios. Critically, the selected tasks must probe the intrinsic capability bottlenecks of existing foundational models while exhibiting robust learnability under few-shot finetuning regimes. This ensures that each incremental learning step represents a meaningful expansion of the model's capability boundaries, accurately simulating the operational demands of scalable generative systems.

Guided by the aforementioned principles, we carefully curate a sequence of six representative tasks encompassing Physical Perception, Entity Manipulation, and Cognitive Reasoning. This composition ensures that the benchmark thoroughly tests the model's capacity to decouple and retain complex, non-overlapping data distributions. The final benchmark evaluates sequential learning across the following distinct image editing tasks:
\begin{itemize}[leftmargin=1.5em, itemsep=0em, topsep=0em]
    \item \textbf{ERP Outpainting:} Spatial reconstruction of panoramic images, requiring the model to contextually synthesize extensive masked regions within Equirectangular Projections (ERP) while maintaining global geometric consistency.
    \item \textbf{Refocus:} Shifting the focal plane or depth of field through localized blur and sharpening while preserving the scene composition.
    \item \textbf{Relighting:} Manipulating scene illumination while preserving structure and maintaining consistent shadows and surface shading.
    \item \textbf{Text Editing:} Locally modifying textual content while preserving font style, layout, and background consistency.
    \item \textbf{Virtual Try-On:} Synthesizing target garments on human subjects while preserving identity under complex poses and occlusions.
    \item \textbf{Causal Reasoning:} Applying causal, physics-grounded transformations that reflect plausible physical or chemical state changes.
\end{itemize}

\noindent \textbf{Data curation.}
CIE-Bench is constructed from high-fidelity image pairs carefully curated for continual image editing tasks.
The raw data is collected from a heterogeneous mixture of existing domain-specific datasets, internet-sourced imagery, and outputs generated by state-of-the-art specialized editing models, ensuring broad coverage of diverse visual distributions and editing scenarios.

\begin{table*}
\begin{center}
\caption{Effectiveness of Adaptive Orthogonal Decoupling (AOD). We compare two constraint strategies, AOD and parameter orthogonality, and report their \textit{Last} and \textit{Imm.} metrics for each task.}
\label{tab_param_orth}
\scalebox{0.9}{
\begin{tabular}{cccccccc|c|c}
\toprule
\multicolumn{2}{c}{Method}  & ERP & Focus & Light & Text & Tryon & Causal & \textbf{Avg.} & \textbf{BWT}\\
\hline
\multicolumn{1}{c}{\multirow{2}{*}{Param. Orth.}} & Imm. & 9.0403 & \textbf{8.5877} & 8.0246 & 9.2295 & \textbf{9.5362} & 9.1702 & 8.9314 & \multirow{2}{*}{-0.5358} \\
 & \cellcolor{gray!15}Last & \cellcolor{gray!15}8.3548 & \cellcolor{gray!15}6.6794 & \cellcolor{gray!15}7.7049 & \cellcolor{gray!15}9.2541 & \cellcolor{gray!15}9.2101 & \cellcolor{gray!15}9.1702 & \cellcolor{gray!15}8.3956 &  \\
\hline
\multicolumn{1}{c}{\multirow{2}{*}{\textbf{AOD}}} & Imm. & \textbf{9.0645} & 8.3588 & \textbf{9.1066} & \textbf{9.7869} & 9.4855 & \textbf{9.4298} & \textbf{9.2054} & \multirow{2}{*}{\textbf{-0.1702}} \\
 & \cellcolor{gray!15}Last & \cellcolor{gray!15}\textbf{8.4274} & \cellcolor{gray!15}\textbf{7.9466} & \cellcolor{gray!15}\textbf{8.1885} & \cellcolor{gray!15}\textbf{9.5738} & \cellcolor{gray!15}\textbf{9.6449} & \cellcolor{gray!15}\textbf{9.4298} & \cellcolor{gray!15}\textbf{8.8685} &  \\
\bottomrule
\end{tabular}
}
\end{center}
\end{table*}

All image pairs are standardized through a unified preprocessing pipeline to maintain consistent resolution, visual quality, and formatting across domains. To guarantee reliable optimization targets, all candidate pairs undergo a rigorous multi-stage manual filtering process. We retain only high-quality instances where the target edit is accurately executed while task-irrelevant regions and structural content remain faithfully preserved. Samples containing artifacts, semantic inconsistencies, excessive distortions, or unintended background modifications are systematically removed.

\renewcommand{\arraystretch}{0.9}
\begin{table}
\caption{Quantitative comparison (\textit{IF}\&\textit{PN}) of different methods on CIE-Bench. In the table, we report the \textit{Last} metric of different methods on each dataset, organized by {\textit{IF}}/{\textit{PN}}.} \vspace{-1mm}
\label{tab_main_IF_PN}
\begin{center}
\scalebox{0.78}{
\begin{tabular}{lc|cccccc|c}
\toprule
Method & Replay & ERP & Focus & Light & Text & Virtual Tryon & Causal & {Avg.}\\
\hline
Zero-shot & - & 3.19/3.87 & 2.79/1.68 & 1.19/1.77 & 3.24/2.65 & 1.17/3.43 & 3.17/3.95 & 2.46/2.89\\
MTFT & - & 3.99/4.98 & 4.56/4.44 & 4.56/4.51 & 4.90/4.89 & 4.97/4.91 & 4.92/4.98 & 4.65/4.78\\
\hline\hline
Sequential FT & - & 3.23/4.13 & 2.72/3.09 & 2.23/2.36 & 4.30/4.04 & 4.72/4.56 & 4.45/4.83 & 3.61/3.84\\
Data Rehearsal & \scalebox{0.90}{\yes} & 3.09/4.85 & 4.56/4.18 & 3.91/3.64 & 4.46/4.46 & 4.78/4.50 & 4.68/4.80 & 4.25/4.40\\

\multicolumn{9}{>{\columncolor{gray!30}}c}{\textcolor{olive!20!black}{$\triangledown$ LoRA Merging}} \\
TIES~\cite{yadav2023ties} & \scalebox{0.90}{\no} & 3.23/4.85 & 2.44/3.11 & 1.88/2.66 & 3.75/3.52 & 2.25/3.80 & 3.67/4.36 & 2.87/3.72\\
DARE~\cite{yu2024language} & \scalebox{0.90}{\no} & 3.20/4.84 & 2.42/2.83 & 1.65/2.44 & 3.63/3.49 & 2.25/3.88 & 3.68/4.43 & 2.80/3.65\\

\multicolumn{9}{>{\columncolor{gray!30}}c}{\textcolor{olive!20!black}{$\triangledown$ Architecture-based}} \\
CL-MoE~\cite{huai2025cl} & \scalebox{0.90}{\no} & 3.25/4.76 & 2.81/3.33 & 2.04/2.57 & \underline{4.11}/3.99 & \underline{4.78}/\underline{4.51} & 4.22/4.53 & 3.53/3.95\\
MoELoRA~\cite{chen2024coin} & \scalebox{0.90}{\no} & 3.02/4.70 & 2.31/2.83 & 1.98/2.36 & 4.02/3.80 & 4.67/4.43 & \textbf{4.70}/\underline{4.85} & 3.45/3.83\\
HiDe-LLaVA~\cite{guo2025hide} & \scalebox{0.90}{\no} & \textbf{3.53}/4.89 & 1.71/2.98 & 2.56/3.22 & 4.07/\underline{4.27} & 1.01/1.28 & 3.51/4.01 & 2.73/3.44\\
SMoLoRA~\cite{wang2025smolora} & \scalebox{0.90}{\no} & 2.89/4.83 & \underline{3.58}/\underline{3.29} & 3.43/3.26 & 4.06/3.86 & 4.36/3.67 & 4.49/4.64 & 3.80/3.93\\

\multicolumn{9}{>{\columncolor{gray!30}}c}{\textcolor{olive!20!black}{$\triangledown$ Regularization-based}} \\
O-LoRA~\cite{wang2023orthogonal} & \scalebox{0.90}{\no} & 3.42/\underline{4.95} & 3.07/3.22 & 3.20/3.07 & 3.87/3.64 & 4.70/4.38 & 4.16/4.54 & 3.74/3.97\\
RegLoRA~\cite{chen2025sefe} & \scalebox{0.90}{\no} & 3.13/4.90 & 3.56/3.01 & \underline{3.87}/\underline{3.52} & 3.81/3.61 & \underline{4.78}/4.29 & 4.15/4.61 & \underline{3.88}/\underline{3.99}\\
KeepLoRA~\cite{luo2026keeplora} & \scalebox{0.90}{\no} & 3.23/4.87 & 1.88/2.47 & 3.29/3.15 & 4.04/3.70 & 4.57/4.24 & 3.99/4.19 & 3.50/3.77\\
\cline{1-9}
{\textbf{ACE-LoRA (ours)}} & \scalebox{0.90}{\no} & \underline{3.43}/\textbf{5.00} & \textbf{4.31}/\textbf{3.63} & \textbf{4.26}/\textbf{3.93} & \textbf{4.80}/\textbf{4.77} & \textbf{4.93}/\textbf{4.72} & \underline{4.56}/\textbf{4.87} & \textbf{4.38}/\textbf{4.49}\\

\bottomrule
\end{tabular}
}
\end{center} \vspace{-5mm}
\end{table}
\renewcommand{\arraystretch}{1.0}
\noindent \textbf{Evaluation protocol.}
Establishing an objective measurement protocol is critical for accurately assessing plasticity and stability in continual learning. Conventional pixel-level metrics inherently fail to capture the complex semantic and structural transformations of advanced image editing. Furthermore, recent MLLM-based evaluators (e.g., ImgJudge~\cite{ye2025imgedit}, VIEScore~\cite{ku2024viescore}) struggle on our specialized downstream tasks despite their success in general domains. These metrics are bottlenecked by two concurrent limitations: (1) capability-reproducibility dilemma, where open-weight models lack fine-grained visual recognition to avoid misjudgments, while capable closed-source APIs compromise reproducibility; and (2) reliance on overly generic prompts that omit implicit domain conventions, rendering evaluators oblivious to common-sense rules essential for fine-grained assessment (see Fig.~\ref{fig-metric})

To circumvent these limitations, we design a customized, reproducible evaluation framework tailored for CIE-Bench. We adopt Qwen3.5-plus~\cite{yang2025qwen3} (open-source version: Qwen3.5-397B-A17B) as our base evaluator, driven by task-specific evaluation prompts formulated with explicit domain priors. Following~\cite{ye2025imgedit}, our pipeline introduces a dual-dimensional assessment strategy, which includes \textit{Instruction Following} (IF, 1-5 ratings) and \textit{Perceptual Naturalness} (PN, 1-5ratings). IF quantifies adherence to target concepts and domain constraints, whereas PN penalizes unnatural artifacts and unintended alterations in non-target regions. We define the Overall Score as the sum of IF and PN, to characterize the overall editing performance. Together, they offer a granular analysis of sequential model performance aligned with real-world standards.

\section{Experiments}
\label{sec:experiments}

\subsection{{Experimental Setup}}
\noindent\textbf{Baseline.}
We adopt Flux2-Klein-9B~\cite{flux-2-2025} as the base model and compare \our\ with representative continual learning methods (including architecture-based and regularization-based methods) and model merging techniques on CIE-Bench. Consistent with prior work, we further include Zero-Shot, Multi-Task Fine-Tuning (MTFT), and Sequential Fine-Tuning (Sequential FT, which sequentially fine-tunes a single LoRA across all tasks), that are commonly regarded as the empirical lower bound, upper bound, and baseline for continual learning methods, respectively. We also report the performance of Data Rehearsal as a reference.

\noindent\textbf{Evaluation Metrics.}
Following prior work, we evaluate continual learning performance using \textit{Last} and \textit{Avg}. \textit{Last} measures the performance on all tasks after the completion of sequential fine-tuning, while \textit{Avg} reports the average performance across tasks during sequential fine-tuning process. In some experiments, we report the \textit{Imm.} and Backward Transfer (BWT). The former refers to the performance evaluated after training on the current task, while the latter quantifies the average degree of forgetting on previously learned tasks during the acquisition of new tasks.
For each task, we follow our evaluation pipeline to assess instruction following (IF), perceptual naturalness (PN), and overall score (OS = IF + PN).
\subsection{{Main Results}}
\begin{figure}
    \centering
    \includegraphics[width=\linewidth]{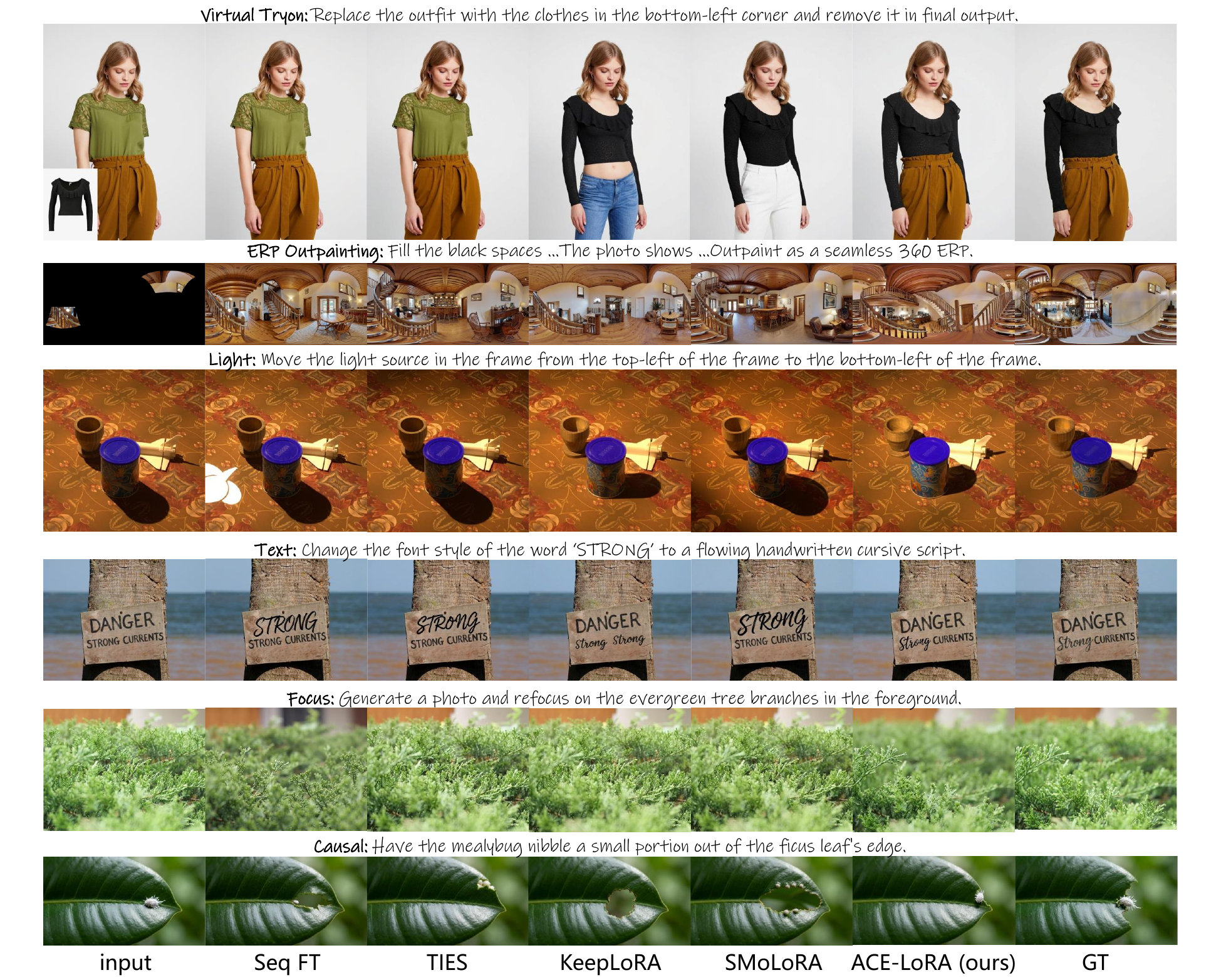} \vspace{-3mm}
    \caption{Qualitative comparison of visual results for different methods on CIE-Bench. For each dataset, we selected a representative example and attached the editing instruction above the image.}
    \label{fig-main_results} \vspace{-0mm}
\end{figure}

\subsubsection{Quantitative Evaluation}

Tab. \ref{tab_main} and Tab. \ref{tab_main_IF_PN} present comprehensive quantitative comparisons of various methods evaluated on CIE-Bench.

\begin{table*}
\begin{center}
\caption{Results of adopting different task orders. In the table, we report the \textit{Last} metric of each task under different task orders. We use the initials to represent different tasks; see Sec. \ref{order} for details.}
\label{tab_order}
\scalebox{0.9}{
\begin{tabular}{ccccccc|c}
\toprule
{Task Order}  & ERP & Focus & Light & Text & Tryon & Causal & \textbf{Avg.}\\
\hline
CVLEFT & 8.0806 & 7.7656 & 7.9583 & 9.6825 & 9.3676 & 8.9464 & 8.6335 \\
FVELTC & 8.1452 & 7.5234 & 8.0081 & 9.2899 & 9.1618 & 9.0797 & 8.5347 \\
EFLTVC & 8.4274 & 7.9466 & 8.1885 & 9.5738 & 9.6449 & 9.4298 & 8.8685 \\
\bottomrule
\end{tabular}
}
\end{center} \vspace{-4mm}
\end{table*}
\noindent \textbf{Inefficacy of Existing Paradigms.} 
As shown in Tab. \ref{tab_main} and Tab. \ref{tab_main_IF_PN}, LoRA-based merging strategies consistently underperform compared to sequential training paradigms, including standard Sequential Fine-tuning. This result suggests that, within the context of continual learning (CL) for image editing, simple weight interpolation across task-specific adaptations is insufficient for achieving effective task decoupling.
Moreover, existing CL methods do not demonstrate substantial gains over Sequential Fine-tuning, and in some cases even lead to performance degradation. This observation suggests that current CL paradigms designed for MLLMs struggle to generalize seamlessly to image editing models. We attribute this discrepancy to the unique characteristics inherent to image editing, particularly the compounded stochasticities introduced by noise and timestep sampling.

\noindent \textbf{Superiority of \our.} 
By introducing dynamic constraints over current training data, \our\ achieves the best performance on CIE-Bench, substantially outperforming existing CL baselines and even surpassing Data Rehearsal in both Avg and Last metrics. Notably, on tasks such as Text and Virtual\_Tryon, \textit{Last} scores of \our\ are comparable to Individual LoRA, indicating effective mitigation of catastrophic forgetting. Furthermore, Tab. \ref{tab_main_IF_PN} shows that \our\ outperforms CL baselines in both Instruction Following (IF) and Perceptual Naturalness (PN), demonstrating its ability to preserve instruction fidelity while maintaining high visual quality under continual fine-tuning.

\subsubsection{Qualitative Analysis}

Fig. \ref{fig-main_results} provides a qualitative comparison of different methods on CIE-Bench. Visualizations reveal that \our\ manifests a distinct superiority over existing CL baselines in both Instruction Following and Perceptual Naturalness on historical tasks.
Across examples from the Focus, Light, and Text tasks, \our\ strictly adheres to the editing prompts, yielding generation outcomes that are highly consistent with the ground truth (GT), whereas baseline methods suffer from severe semantic drift or instruction neglect due to catastrophic forgetting.
Furthermore, \our\ effectively reserves the structural constraints of historical tasks while maintaining image naturalness. For instance, in ERP Outpainting, \our\ ensures seamless alignment between the left and right boundaries, achieving global continuity that is often broken by baseline methods. In Virtual Try-on, \our\ precisely restricts edits to the target garment and preserves other regions, whereas baselines tend to introduce unintended modifications beyond the specified area.

\subsection{Model Analysis}
\label{order}

\noindent \textbf{Effectiveness of Adaptive Orthogonal Decoupling.}
Tab. \ref{tab_param_orth} compares AOD with parameter-space orthogonal constraints. Overall, AOD consistently outperforms parameter orthogonalization across both plasticity and stability metrics, with particularly significant gains in stability.
This is due to AOD’s data-dependent modeling of interference, which identifies and suppresses only task-conflicting directions while preserving shared optimization directions beneficial for both historical and current tasks. In contrast, parameter-space orthogonalization enforces uniform constraints on all parameter directions, which may unnecessarily restrict shared knowledge and thus limit plasticity.

\noindent \textbf{Strategies for historical information compression}
\begin{table}[t]
  \centering
  \small
  \caption{Effectiveness of different strategies for history information compression (Imm./Last/BWT).}
  \label{tab_history}
  \scalebox{0.86}{%
    \begin{tabular}{cccccccc|c|c}
      \toprule
      \multicolumn{2}{c}{Method}  & ERP & Focus & Light & Text & Tryon & Causal & \textbf{Avg.} & \textbf{BWT}\\
      \hline
      \multicolumn{1}{c}{\multirow{2}{*}{Random}} & Imd. & 8.8226 & 8.3125 & 8.7916 & 8.2750 & 9.2147 & 8.6160 & 8.6721 & \multirow{2}{*}{-0.2802} \\
       & \cellcolor{gray!15}Last & \cellcolor{gray!15}8.3548 & \cellcolor{gray!15}7.9375 & \cellcolor{gray!15}7.5416 & \cellcolor{gray!15}8.7250 & \cellcolor{gray!15}9.1764 & \cellcolor{gray!15}8.6160 & \cellcolor{gray!15}8.3919 &  \\
      \hline
      \multicolumn{1}{c}{\multirow{2}{*}{Summation}} & Imd. & 8.1290 & 7.8281 & 8.6583 & 8.2750 & 9.0809 & 9.1785 & 8.5250 & \multirow{2}{*}{-0.3030} \\
       & \cellcolor{gray!15}Last & \cellcolor{gray!15}7.8065 & \cellcolor{gray!15}6.3359 & \cellcolor{gray!15}7.8750 & \cellcolor{gray!15}9.0333 & \cellcolor{gray!15}9.1029 & \cellcolor{gray!15}9.1785 & \cellcolor{gray!15}8.2220 &  \\
      \hline
      \multicolumn{1}{c}{\multirow{2}{*}{SVD}} & Imd. & \textbf{9.0645} & \textbf{8.3588} & \textbf{9.1066} & \textbf{9.7869} & \textbf{9.4855} & \textbf{9.4298} & \textbf{9.2054} & \multirow{2}{*}{\textbf{-0.1702}} \\
       & \cellcolor{gray!15}Last & \cellcolor{gray!15}\textbf{8.4274} & \cellcolor{gray!15}\textbf{7.9466} & \cellcolor{gray!15}\textbf{8.1885} & \cellcolor{gray!15}\textbf{9.5738} & \cellcolor{gray!15}\textbf{9.6449} & \cellcolor{gray!15}\textbf{9.4298} & \cellcolor{gray!15}\textbf{8.8685} &  \\
      \bottomrule
    \end{tabular}%
  }
\vspace{-2mm}
\end{table}

Tab. \ref{tab_history} compares random sampling (randomly sampling a single historical task at each training step), summation (summing $LoRA_A$ and $LoRA_B$ over historical tasks), and SVD-based compression. Overall, SVD performs best across all metrics, achieving the highest average scores in \textit{Imm.} and \textit{Last} as well as the least negative BWT, indicating both more effective mitigation of forgetting and improved plasticity.
The advantage of SVD lies in its ability to preserve the principal directions of historical knowledge while filtering redundancy, enabling more accurate interference estimation and maintaining strong model plasticity.

\noindent\textbf{Sensitivity to task order.}
We evaluate \our\ under different task sequences (Tab.~\ref{tab_order}) and observe stable performance across orders (Avg: 8.53--8.87), indicating low sensitivity to task permutations.


\noindent\textbf{Validity of MLLM-as-a-Judge.} Human study details (Tab.~\ref{tab_absolute}, Tab.~\ref{tab_relative}) are deferred to Appendix~\ref{sec:appendix-human-study}.

\vspace{-0.6em}
\section{Conclusion}
In this work, we propose \our, a dynamic regularization framework for continual image editing under parameter-efficient fine-tuning. \our\ mitigates catastrophic forgetting via Adaptive Orthogonal Decoupling and remains scalable with rank-invariant historical information compression. We also introduce CIE-Bench, the first comprehensive benchmark for continual image editing with a standardized evaluation protocol. Extensive experiments demonstrate consistent improvements over strong baselines. Limitations are detailed in the Appendix.

\noindent \textbf{Limitations.}
ACE-LoRA introduces additional training constraints, increasing training time to about $1.5\times$ that of sequential finetuning, comparable to other regularization-based methods. Like other LoRA-based continual learning approaches, its scalability may be limited as task-specific adapters accumulate across tasks.


{
\small
\bibliographystyle{plain}
\bibliography{main}
}

\newpage
\renewcommand\thesection{\Alph{section}} 
\clearpage

\section*{Appendix}

\setcounter{section}{0}
\section{Related Work}
\noindent\textbf{Diffusion-Based Image Editing.}
Large-scale diffusion models~\cite{rombach2022high, podell2023sdxl, esser2024scaling, labs2025flux, peebles2023scalable} have demonstrated remarkable success in synthesizing high-fidelity and semantically complex images from textual prompts. Leveraging these powerful generative priors for image editing, zero-shot approaches formulate editing as controlled trajectory manipulation in the generative process. Specifically, SDEdit~\cite{meng2021sdedit} performs zero-shot editing by injecting stochastic noise into input images and subsequently reversing the diffusion process conditioned on target prompts. To better preserve fine-grained structural consistency, Prompt-to-Prompt~\cite{hertz2022prompt} replaces cross-attention maps associated with source text tokens during latent denoising steps. 
Supervised approaches like InstructPix2Pix~\cite{brooks2023instructpix2pix} fine-tune diffusion models on large-scale instruction–image pairs, enabling direct image editing in a single forward pass. For efficient adaptation to new concepts, low-rank adaptation techniques such as LoRA~\cite{hu2022lora} constrain parameter updates within low-rank subspaces, while methods like CustomDiffusion~\cite{kumari2023multi} further optimize these adapters for rapid customization of target concepts. However, sequential fine-tuning of low-rank modules in dynamic editing scenarios can lead to unintended interference in the learned subspaces, effectively overwriting previously learned representations. This uncontrolled rank-space interference may result in progressive degradation of historical knowledge, motivating the need for explicit mechanisms to stabilize parameter updates in continual editing pipelines.

\noindent\textbf{Continual Learning in Foundation Models.}
The central objective of continual learning (CL) is to mitigate catastrophic forgetting in sequential task learning. Current CL methods can be broadly categorized into architecture-, rehearsal-, and regularization-based approaches.
Architecture-based methods preserve task-specific knowledge by allocating dedicated model components. For instance, CODA-Prompt~\cite{smith2023coda} maintains an expandable pool of learnable prompts, while MoE-Adapters~\cite{yu2024boosting} route representations through specialized expert modules conditioned on the input task.
Rehearsal-based methods alleviate forgetting by retaining or approximating past data distributions. For example, DDGR~\cite{gao2023ddgr} leverages pre-trained diffusion models to synthesize representative historical samples, while memory-efficient approaches such as PCR~\cite{lin2023pcr} maintain compact exemplar sets and employ proxy-based contrastive learning to preserve past knowledge.
Regularization-based methods constrain parameter updates to reduce interference with previously learned tasks. OGD~\cite{farajtabar2020orthogonal} enforces orthogonality between current updates and the gradient subspace of prior tasks. Similarly, O-LoRA~\cite{wang2023orthogonal} mitigates interference by learning task-specific low-rank adapters under orthogonality constraints.

\noindent\textbf{Continual Learning for Generative Models.}
Preserving generative capabilities under sequential task shifts requires carefully balancing plasticity and stability between historical knowledge and new adaptations. Lifelong GAN~\cite{zhai2019lifelong} addresses this issue by applying knowledge distillation on intermediate representations of a frozen generator to maintain consistency. In low-data regimes, LFS-GAN~\cite{seo2023lfs} restricts parameter updates to normalization layers while preserving source-domain fidelity through contrastive patch-level objectives.
In diffusion-based generative models, Null-text Inversion~\cite{mokady2023null} avoids direct parameter updates by optimizing unconditional text embeddings, enabling faithful reconstruction of real images without modifying core network weights. To systematically evaluate sequential adaptation in text-to-image models, T2I-ConBench~\cite{huang2025t2i} introduces metrics for quantifying semantic degradation across consecutively fine-tuned concepts.

Despite these advances, continual learning for image editing remains underexplored. To address this gap, we propose \our, a regularization-based framework tailored for continual image editing, and introduce the first comprehensive benchmark along with a standardized evaluation protocol for assessing continual image editing performance.

\section{More Details of CIE-Bench}
\subsection{More Examples}
\begin{figure}
    \centering
    \includegraphics[width=\linewidth]{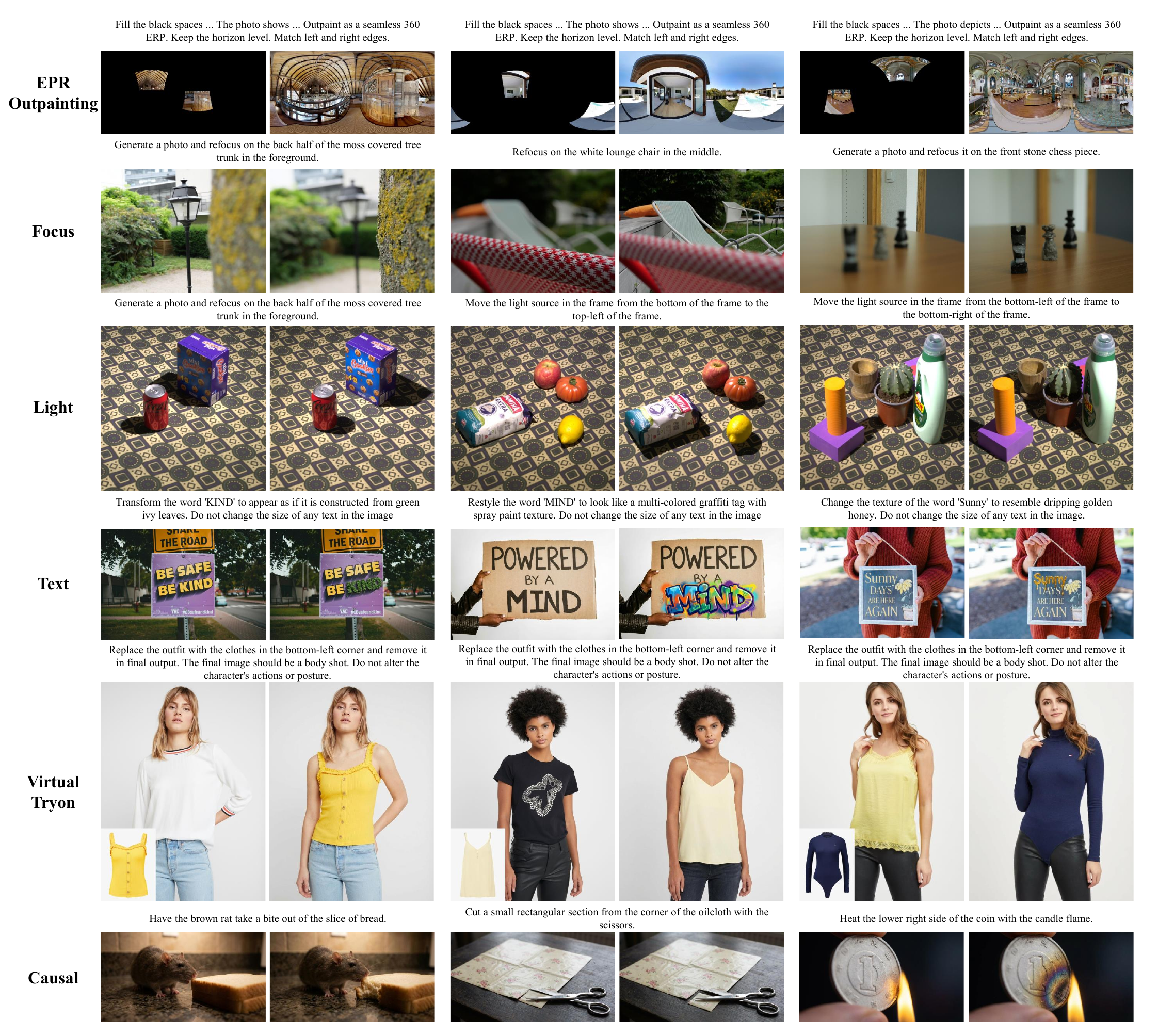}
    \caption{Visualization of CIIE-Bench, which consists of six sub-tasks: ERP Outpainting, Refocus, Relighting, Text Editing, Virtual Try-on, and Causal Reasoning}
    \label{fig:examples}
\end{figure}

We present more examples from CIE-Bench in Fig. \ref{fig:examples}. For each task, we provide three examples, each consisting of the original image (\textbf{left}), the edited image (\textbf{right}), and the editing instruction displayed above the images.

\section{More Analysis}
\subsection{Implementation Details and Experiments Compute Resources.}
We use Flux2-Klein-9B~\cite{flux-2-2025} as the base model, set the LoRA rank of all methods to 48, train for 10 epochs on each task, and use a learning rate of 1e-4.
We use four H20 GPUs for training, each equipped with 96GB of memory, and the server has a total of 1TB of RAM. For our method, the training time per task ranges from 0.5 to 2 hours, and the total training time for the entire benchmark is approximately 8 hours.

\section{Human Study for MLLM-as-a-Judge}
\label{sec:appendix-human-study}
\begin{table}[t]
  \centering
  \small
  \setlength{\tabcolsep}{3pt}
  \renewcommand{\arraystretch}{0.92}
  \caption{Alignment ratio of MLLM evaluator with human preferences in absolute scores.}
  \label{tab_absolute}
  \begin{tabular}{lcccccc|c}
    \toprule
    {Score}  & ERP & Focus & Light & Text & Tryon & Causal & \textbf{Avg.}\\
    \hline
    Instruction Following & 0.4762 & 0.6571 & 0.8190 & 0.8000 & 0.9143 & 0.7619 & 0.7381 \\
    Perceptual Naturalness & 0.8667 & 0.7143 & 0.9476 & 0.8762 & 0.6476 & 0.8095 & 0.8103 \\
    Overall Score & 0.6476 & 0.6095 & 0.8667 & 0.7048 & 0.6667 & 0.7333 & 0.7048 \\
    \bottomrule
  \end{tabular}
\end{table}

\begin{table}[t]
  \centering
  \small
  \setlength{\tabcolsep}{3pt}
  \renewcommand{\arraystretch}{0.92}
  \caption{Alignment ratio of MLLM evaluator with human preferences in relative rankings.}
  \label{tab_relative}
  \begin{tabular}{lcccccc|c}
    \toprule
    {Score}  & ERP & Focus & Light & Text & Tryon & Causal & \textbf{Avg.}\\
    \hline
    Instruction Following & 0.6800 & 0.7708 & 0.9286 & 0.9811 & 0.9190 & 0.6667 & 0.8244 \\
    Perceptual Naturalness & 0.9600 & 0.7778 & 0.9655 & 0.9333 & 0.5625 & 0.8889 & 0.8480 \\
    Overall Score & 0.8293 & 0.7126 & 0.9508 & 0.9200 & 0.8621 & 0.7021 & 0.8295 \\
    \bottomrule
  \end{tabular}
\end{table}
To validate the effectiveness of the MLLM-as-a-Judge framework, we conduct a human study from both absolute and relative perspectives. We select 90 editing samples, with 15 samples for each task, covering diverse editing quality levels.

For absolute evaluation, following~\cite{ye2025imgedit}, we compare the scores assigned by the MLLM evaluator with those given by human experts for each edited sample. A match is counted when the absolute score difference is within 1.

For relative evaluation, we assess pairwise ranking agreement, i.e., whether the MLLM evaluator and human experts agree on which sample in each pair exhibits higher editing quality. This setting is motivated by potential discrepancies in absolute score calibration, particularly for partially successful edits.

\end{document}